\documentclass[runningheads]{llncs}

\usepackage{graphicx}
\usepackage{cite}
\usepackage{amsmath,amssymb,amsfonts}
\usepackage{textcomp}
\usepackage{xcolor}
\usepackage{times}
\usepackage{multirow}
\usepackage{listings}
\usepackage{siunitx}
\usepackage{booktabs}
\usepackage{amsmath}
\usepackage[ruled,vlined]{algorithm2e}
\usepackage{array}
\usepackage{hyperref}
\usepackage{ulem}
\usepackage{subfig}
\usepackage{epstopdf}
\usepackage{epsfig}

\let\ACMmaketitle=\maketitle
\renewcommand{\maketitle}{\begingroup\let\footnote=\thanks \ACMmaketitle\endgroup}

\def\BibTeX{{\rm B\kern-.05em{\sc i\kern-.025em b}\kern-.08em
		T\kern-.1667em\lower.7ex\hbox{E}\kern-.125emX}}

\begin{document}

\title{Feature Extraction Functions for Neural Logic Rule Learning}

\author{Shashank Gupta \ \ \ \ \ \  \ \  \ \ \ \ \ \  \ Antonio Robles-Kelly\\ 
\vspace{0.2cm}
Mohamed Reda Bouadjenek}

\authorrunning{Shashank Gupta et al.}

\institute{School of Information Technology, Deakin University, Waurn Ponds Campus,\\ Geelong, VIC 3216, Australia}

\maketitle

\begin{abstract}

 Combining symbolic human knowledge with neural networks provides a rule-based ante-hoc explanation  of the output. 
 In this paper, we propose feature extracting functions for integrating human knowledge abstracted as logic rules into the predictive behaviour of a neural network. 
 These functions are embodied as programming functions, which represent the applicable domain knowledge as a set of logical instructions and provide a modified distribution of independent features on input data.
 Unlike other existing neural logic approaches, the programmatic nature of these functions implies that they do not require any kind of special mathematical encoding, which makes our method very general and flexible in nature. We illustrate the performance of our approach for sentiment classification and compare our results to those obtained using two baselines.

\end{abstract}

\section{Introduction}
\label{sct:01}
Deep Neural Networks tend to suffer from the Black Box problem, mainly because their training is often purely data-driven, with no direct or indirect human intervention~ \cite{arun:2020}. As a result, the interpretation of the input-output mapping is often challenging, if not almost intractable. Moreover, they do not have an inherent representation of causality or logical rule application.
Indeed, previous work has shown that supervision purely in the form of data can lead a model to learn some unwanted patterns and provide misleading and incorrect predictions \cite{Szegedy:2014,Nguyen:2015}.
These drawbacks hinder their applications in a wide range of domains such as cyber-security, healthcare, food safety, power generation and environmental management, which require a level of trust or confidence associated with the output of the network~\cite{Ribeiro:2016}.

A common approach to make the predictions of a Neural Network explainable is to encode the intended rules or patterns derived from human domain knowledge in its trainable parameters~\cite{vilone:2020}.  
This can be viewed as the process of combining structured logical knowledge representing high-level cognition with neural systems~\cite{Garcez:2002}. 
Indeed, logic rules provide a way to represent human knowledge in a structured format. 
However, logic rules need to be translated from natural language to logical representations. 
Moreover, they  require a suitable encoding format, which is not a straightforward task because in most cases this encoding is application-specific. 

\begin{figure}[!t]
\begin{center}
\includegraphics[width = \textwidth]{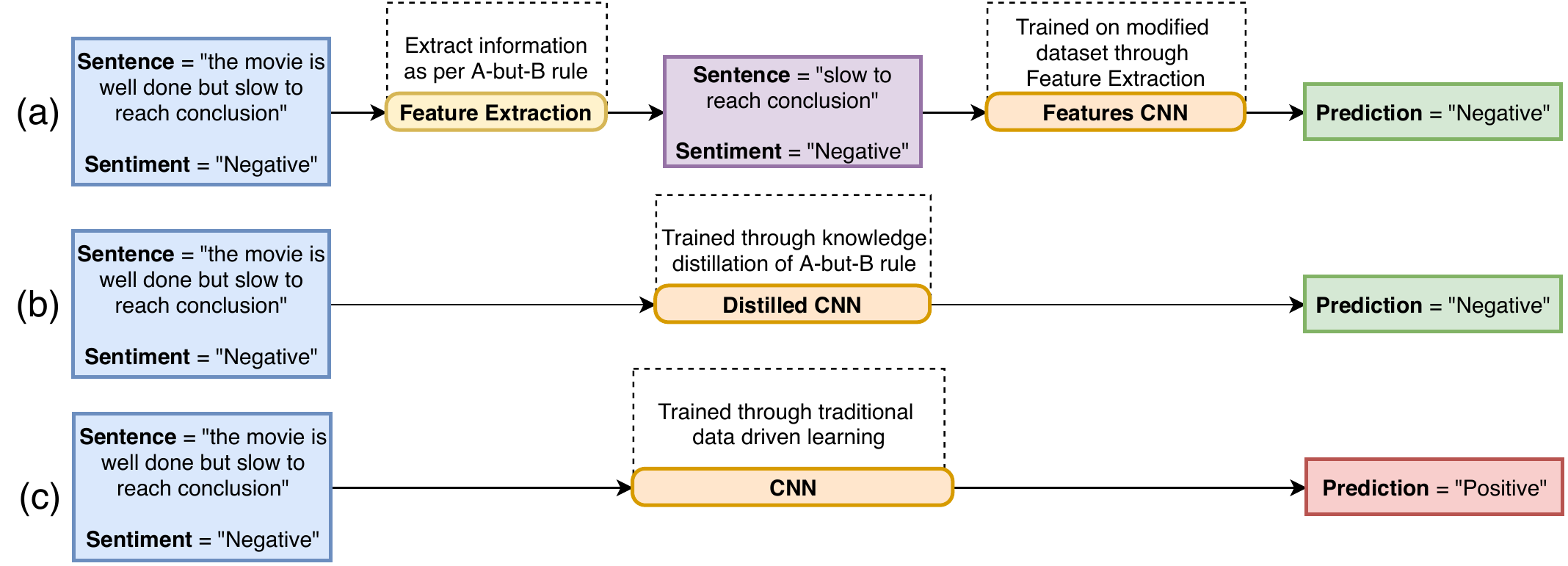}
\end{center}
\caption{(a) Overview of our proposed approach, which abstracts domain knowledge into the sentiment prediction of a neural network using feature extraction on the input data instead of distillation (bottom-way prediction). This achieves an ante-hoc rule based explanation of Neural Network inferential process as compared to (b), a distillation approach (middle-way prediction), which encodes knowledge into the network parameters or to (c), a straight application of a CNN to the sentence-sentiment tuple (top-way prediction) devoid of neural logic.}
\label{fig:Teaser_Figure}
\end{figure}

One way to efficiently encode human knowledge abstracted as first order logic rules into the parameters of a neural network is to use the iterative-knowledge distillation method~\cite{Hu:2019}, a process summarized in Figure~\ref{fig:f1}. 
Briefly, iterative-knowledge distillation consists of representing structured human knowledge as a set of declarative first-order logic rules using soft-logic \cite{Bach:2015}, then, encoding these rules into the parameters of the network via indirect supervision making use of knowledge distillation \cite{Hinton:2015} at each training iteration. 
However, while iterative knowledge distillation
makes the network to learn from both data and rules, we find that it implicitly makes an assumption of knowledge to remain static and true for every data point in the data set. 
Also, it imprints the knowledge into the network parameters permanently through distillation and do not provide any mechanism to accommodate for any change in the existing rules or addition of new ones. 
Thus, updating the rules requires to re-train the whole network. 
This can sometimes lead to a decrease in performance as shown in our experimental results.

To overcome the aforementioned issues, we propose to construct feature-extracting functions instead of logic rules from human knowledge as summarized in Figure~\ref{fig:Teaser_Figure}. 
These functions are analogous to decision rules~\cite{Ratner:2017} but modified to provide supervision similar to logic rules~\cite{Hu:2019}. 
They are directly applied on the data so as to transfer the human knowledge into a distribution of the input data and influence the output of the network. 
We do this by viewing each function as a mini-batch processing step during each iteration. 
Since the functions are applied directly to the data, we do not need to compute the probability distributions nor construct a teacher network. 
This effectively reduces the complexity of our method. 
Also, these feature-extracting functions can be modified at any time during the training process, thus providing a lot of flexibility in adapting to qualitative and quantitative characteristics of the data under consideration. 
This is consistent with the well known properties of feature-extracting functions to express natural language~\cite{lewis:92}, exploiting these traits for the training of deep networks to provide a more direct nature of supervision based upon the input data. 
Our method is quite general in nature, being a flexible manner of providing human knowledge supervision to the network, hence, it can be applied to tasks beyond Natural Language Processing. 

\section{Related Work}

A lot of research has been done in the past few years for incorporating domain knowledge about a problem into machine learning models \cite{Tran:2017,Hu:2019,Ganchev:2010,Tasker:2003,Liang:2018}. These methods essentially use knowledge represented in logical and/or symbolic form to construct posterior constraints on the model prediction and train the model to capture those constraints. Iterative Knowledge Distillation \cite{Hu:2019} sets itself apart from other neural symbolic methods as it provides a very flexible framework for integrating knowledge represented as first order logic rules with  general purpose neural networks such as CNNs and RNNs. 

A recent paper \cite{krishna:2018} gives a detailed analysis on the methodology used in \cite{Hu:2019}, comparing its performance to other neural symbolic methods and arguing that it is not very effective in transferring knowledge to the neural network model (student network).
Our work is consistent with this finding, achieving better performance by representing knowledge purely in terms of data, which is directly given as input for training a neural network. 
Moreover, since we have used the same data sets as those in \cite{Hu:2019}, we employ an identical type of supervision as that used for the iterative knowledge distillation.

Finally, the authors in~\cite{krishna:2018} suggest using a deep contextualized word representation model such as ELMo (Embeddings from Language Models~\cite{Peters:2018}) and feed the embedding to the neural network to better capture the rule knowledge. However, this still fails to accommodate the dynamic nature of rules acquired from domain knowledge and its only limited to Language-related tasks.

\section{Feature-Extracting Functions and Neural Logic}
\begin{figure*}[!t]
\centering
\subfloat[Iterative Rule-knowledge Distillation]{\includegraphics[width = 0.45\textwidth]{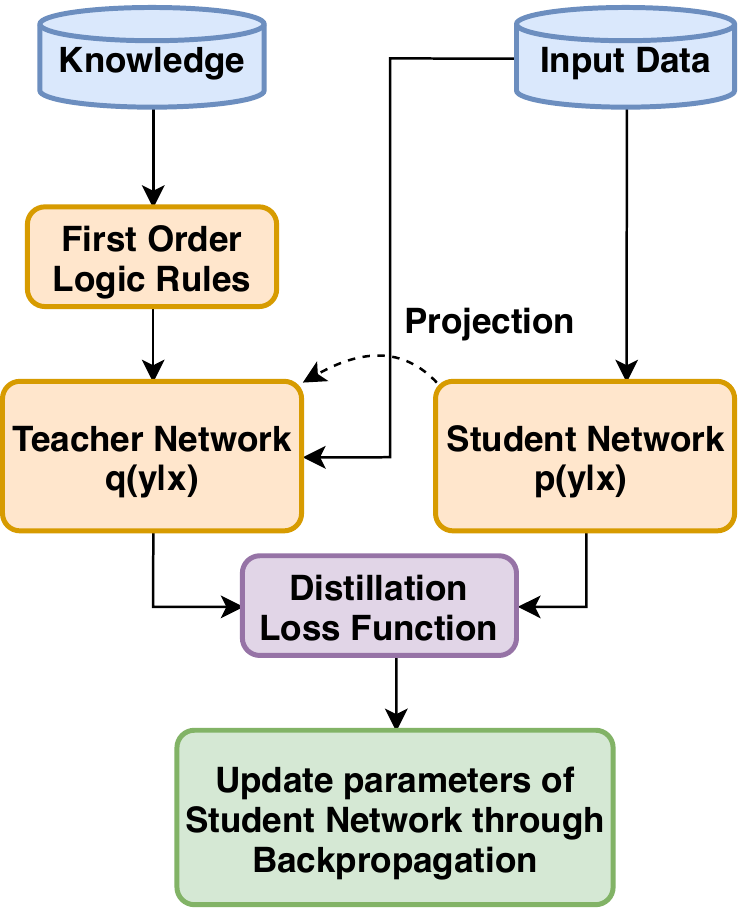}\label{fig:f1}}
\hfill
\subfloat[Feature Extracting Functions]{\includegraphics[width = 0.45\textwidth]{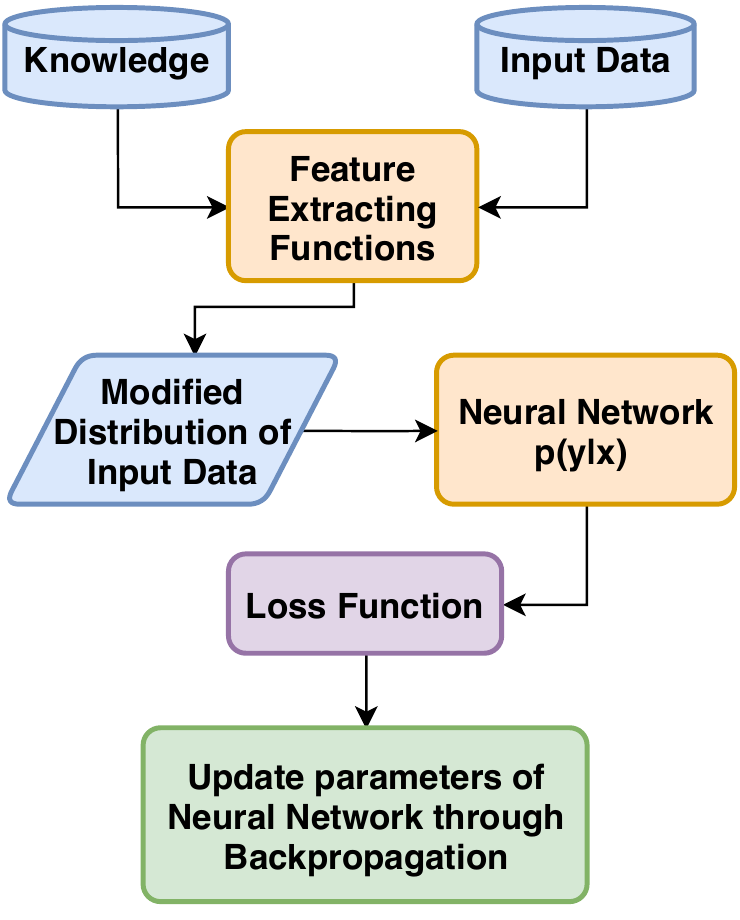}\label{fig:f2}}
\caption{(a) shows an overview of the iterative knowledge distillation framework in \cite{Hu:2019}; (b)  shows an overview of the  method we propose.}
\end{figure*}

In our approach, we develop feature-extracting functions from human knowledge instead of constructing logic rules, which are expressed as programming functions and take the data instances in terms of independent features as input.
They enforce the knowledge directly upon the neural network during training. This eliminates the need for constructing a teacher network and provide the flexibility to allow these functions to be applied either during the training process or during the pre-processing phase of the data. Figure \ref{fig:f2} summarizes our approach.

\subsection{Distillation vs Feature Extraction}
In iterative distillation, a parametric baseline neural network is used as a ``student'', which needs to be provided with logical knowledge by a non-parametric ``teacher'' network. The teacher network is a projection of the student network over a regularized sub-space whereby the training data is constrained by logical rules.
These logic rules are encoded using soft-logic \cite{Bach:2015} for the sake of constructing soft-boundaries and for calculating rule-regularized distributions. 
Thus, the training data comprises a set $D = {\{ (x_n,y_n) \}}_{n=1}^N$ of N tuples $(x_i,y_i)$, where $x_i$ is an input instance (an independent variable or a set of independent variables) and $y$ is the corresponding target. 
The set of logic rules are expressed as $ R = {\{ (R_l,\lambda_l) \}}_{l=1}^L$ where $ R_l $ is the $l^{th}$ rule constructed from human knowledge over $D$ and  $\lambda_l$ is the corresponding confidence value. A logic rule can be made up of several conditions or logic expressions. Each logic expression when instantiated on $D$ produces a set of groundings as $\{ (r_{l_g}(D)) \}_{g=1}^{G_l}$ and thus, represent a rule as a set of ground expressions on $D$ where each $r_{l_g}$ is $g^{th}$ grounding of the $l^{th}$ rule. The combined set of $D$ and $R$ is called \textit{learning resources}.

For example, consider a set of movie reviews in which $x$ comprises a set of tokens and the target $y$ represents the sentiment value (0 for negative and 1 for positive reviews). From human knowledge, we know that, if a sentence has a syntactic structure of ``A-but-B'', then the sentiment of the sentence should be consistent with that of ``B'' component. 
Therefore, we can express the ``A-but-B'' statement as a logic rule stated as $R_1$ with an assumption that at least one ground expression will evaluate to ``True'' ($\lambda_1 = 1$). To encode this formally, we  define a Boolean random variable $r_{l_g}(x,y)$ = ``has an \textit{A-but-B} structure'', then apply an expectation operator on it to calculate sets of valid distributions in $D$, which will be further used to construct a ``teacher network''. This process is complex and time consuming which is not applicable to different types of datasets from different domains.

To tackle this drawback, our method combines the input and human knowledge to provide a pre-processed data set, which can be used for training the neural network. 
For the sake of consistency, we denote the input data  $D=\{(x_n,y_n) \}_{n=1}^N$ as a set of $N$ tuples $(x,y)$, where $x_i$ is a set of input independent variables and its corresponding target $y_i$, and the human knowledge $F = {\{ (F_l(D) \}}_{l=1}^L$ as a set of $L$ feature extracting functions, which are applied on $D$. 
Revisiting the previous example, instead of using soft-logic using auxiliary random variables, for the ``A-but-B'' rule we write a function $F_l = A-but-B(x,y)$, which outputs $(x*,y)$, where $x*$ has only 'B' features to be is consistent with $\lambda_l = 1$ as presented above.

\begin{algorithm}[!t]
\SetAlgoLined
 \textbf{Input:} The training batch set $ D = {\{ (x_n,y_n) \}}_{n=1}^N$, \\
 \quad \quad \quad The functions set $F = {\{ (F_l(D) \}}_{l=1}^L$ \\
 Initialize the neural network parameters \textbf{$\theta$} \\
 \While{Iteration}{
  1: Calculate $ D^* = {\{ (x_n^*,y_n) \}}_{n=1}^N$\\
  2: Calculate the probability distribution $p_\theta(Y|X^*)$\\
  3: Update the parameters \textbf{$\theta$} using objective function in Eq.(2)
 }
 \textbf{Output:} Trained neural network 
 \caption{Training process.}
 \label{algo1}
\end{algorithm}

\subsection{Feature-Extracting Functions}
Consider the conditional probability distribution $p_\theta(y_i|x_i)$ with parameter set $\theta$ as the softmax output of a Neural Network. Here, inspired by the labeling functions used by Ratner {\it et al.} \cite{Ratner:2017}, we use the input instance $x_i$ to compute a post-processed instance $x_i^*$.
We can view the post-processed instance $x_i^*$ as an explicit representation of the domain knowledge, expressed in the rule under consideration and mapped onto the input instance $x_i$. This is an important observation since it hints at a minimisation problem on the cumulative output on the feature extracting functions so as to obtain the parameter set $\theta$ which can be expressed formally as follows:
\begin{equation} \label{eq3}
\begin{split}
\theta & = \arg\min_{\theta\in\Theta}\frac{1}{N}\Sigma_{n=1}^{N}L(y_n,p_\theta(Y|X^*))
\end{split}
\end{equation}

\noindent where $L(\cdot)$ is the loss function of choice and $p_\theta(Y|X)$ is the conditional probability distribution of the target set $Y$ given the set $X^*$ of all the post-processed instances $x_i^*$. Since the information is purely present in the modified feature-set, the feature extracting functions become a post-processed input data for the network.

The treatment above also has the advantage of ease of implementation. We summarise the training and testing process of our method in Algorithm~\ref{algo1}.
Note that at each training iteration, we calculate the post-processed data set $D^* = {\{ (x_n^*,y_n) \}}_{n=1}^N$ using the feature extracting functions $F_l \in F$ as applied on the input batch $ D = {\{ (x_n,y_n) \}}_{n=1}^N$. These are passed  to the neural network to calculate the conditional probability $p_\theta(y_i|x_i^*)$ for each $(x_i^*,y_i) \in D^*$.

\section{Experiments}
We performed sentence-level binary sentiment classification and compared our method (CNN-F) with a baseline network (CNN) devoid of knowledge support and its knowledge distilled version (CNN-rule) created from Iterative knowledge Distillation \cite{Hu:2019}. We used the same convolutional neural network architecture proposed in~\cite{Kim:2014} employing it's ``non-static'' version with the exact same configuration as that presented by the authors. 
We have compared our method CNN-F against the non-static version of the CNN in~\cite{Kim:2014} as published by the authors and the CNN-rule in~\cite{Hu:2019}, which is a knowledge distilled version of CNN.  
Also, we have initialised word vectors using word2vec~\cite{Mikolov:2014} and used fine-tuning, training the neural network using stochastic gradient descent (SGD) with the AdaDelta optimizer~\cite{Zeiler:2012}.

Since contrasting senses are hard to capture, we define a linguistically motivated rule called ``A-but-B'' rule akin to that in~\cite{Hu:2019}. It  states that if a sentence has an ``A-but-B'' syntactic structure, the sentiment of the whole sentence will be consistent with the sentiment of it's ``B'' component. 
For example, for the sentence S =  ``you can taste it , but there 's no fizz'', its sentiment is decided by only the sentiment of its B component = ``there 's no fizz''. From this rule, we can define a feature-extracting function $F_1 = A-but-B(x,y)$ on set $D$ which takes the input pair of sentence-label $(x,y)$ and outputs $(x*,y)$ where $x*$ is corresponding features of ``B''.

We evaluate our method on three public data-sets: 
\begin{enumerate}
    \item The Stanford sentiment tree bank dataset (SST2)~\cite{Socher:2014}, which contains 2 classes (negative and positive), and 6,920/872/1,821 sentences in the train/dev/test sets respectively. Following~\cite{Kim:2014}, we train the models on both, sentences and phrases. 

    \item The movie review one (MR) introduced in~\cite{pang-lee:2005}. This data set consists of 10,662 one-sentence movie reviews with negative or positive sentiments. 

    \item The customer reviews of various products data set (CR) presented in~\cite{Hu-and-Liu:2004}, which contains 2 classes and 3,775 instances\footnote{As we present our method as an alternative to the iterative-knowledge distillation \cite{Hu:2019}, a direct comparison was necessary in terms of results and thus, we adopted the same methodology to produce results as in \cite{Hu:2019}. The authors in \cite{Hu:2019} also employ 10-fold cross validation for the MR and CR data sets}.  

\end{enumerate}

We also evaluate our method only on the sentences containing ``A-but-B'' structure in the test sets of all three data sets under study to show that better performance of our method CNN-F on the whole test set is indeed attributed to the better performance on sentences having ``A-but-B'' structure. SST2 test set has a total of 1,821 instances out of which 210 instances exhibit the ``A-but-B'' structure. For MR data, it has a total of 10,662 instances out of which 1603 instances are found to have ``A-but-B'' structures. Finally, the CR data set has a total of 3,775 instances out of which 413 instances contain sentences with ``A-but-B'' structures. For the MR and CR dataset, we use nested 10-fold cross validation and report mean$\pm$ 95\% confidence interval for all performance metrics over the ten trails corresponding to the 10-fold cross validation. For these results, we have used the models of CNN, CNN-rule and CNN-F trained using the whole data sets.

\begin{figure}[!t]
\centering
\includegraphics[width = 0.9\textwidth]{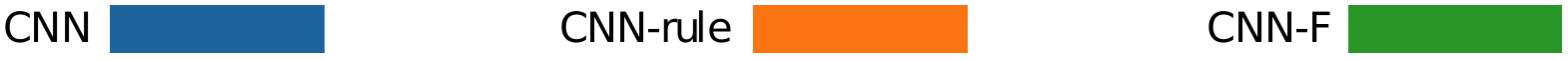}
\subfloat[SST2]{\includegraphics[width = 0.33\textwidth]{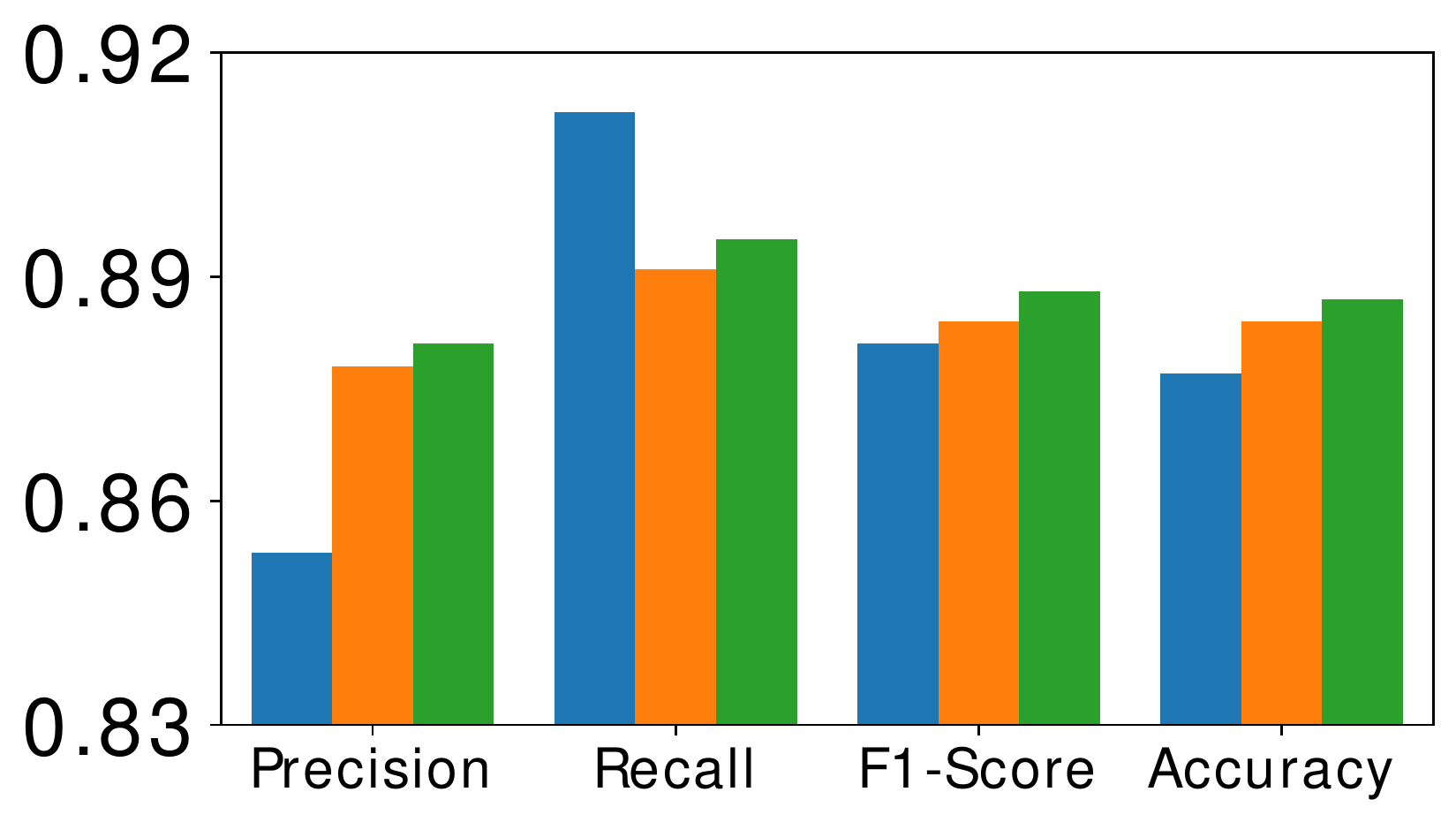}}
\subfloat[MR]{\includegraphics[width = 0.33\textwidth]{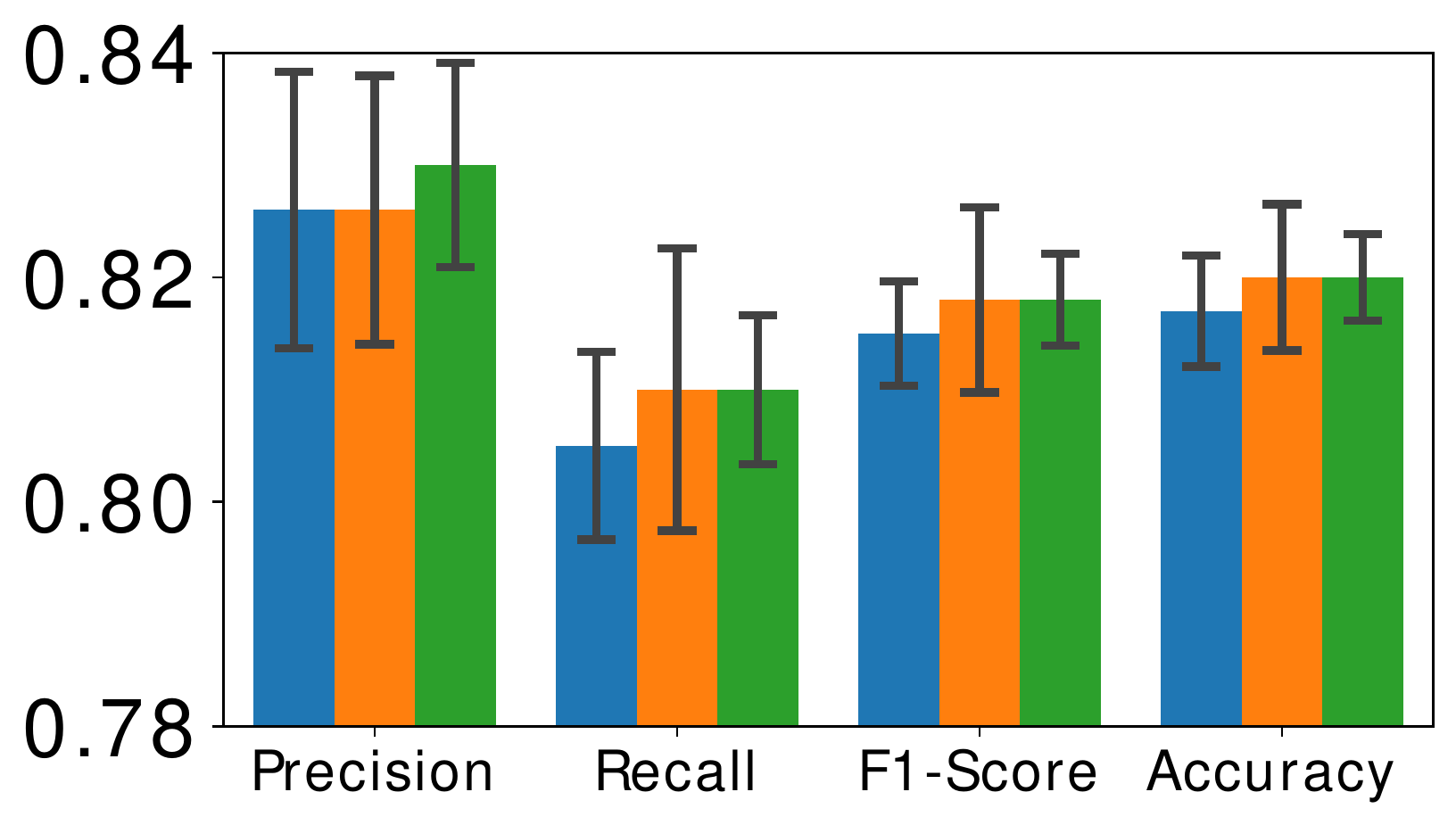}}
\subfloat[CR]{\includegraphics[width = 0.33\textwidth]{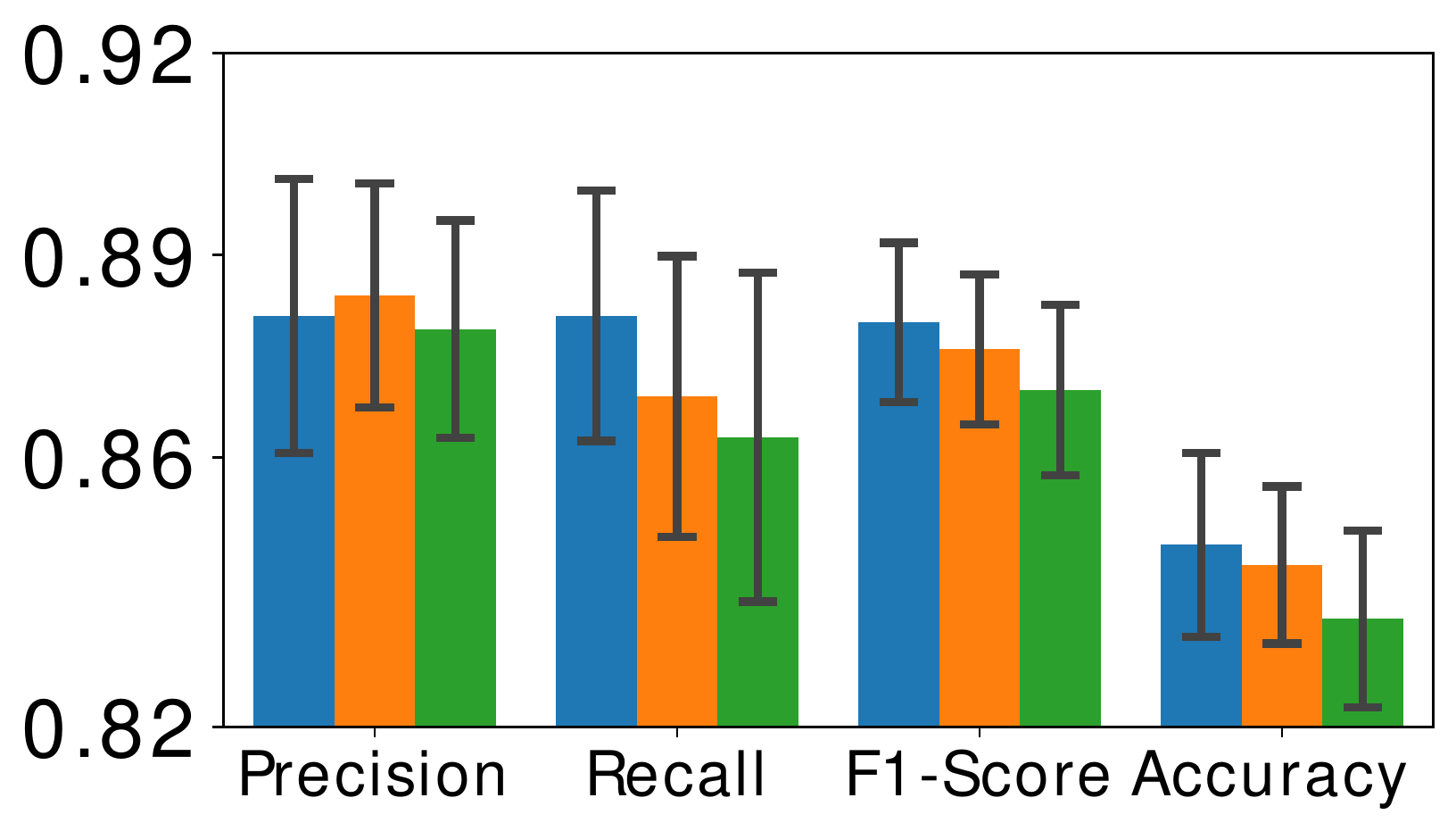}}
\caption{Performance obtained using our method (CNN-F), the method in \cite{Hu:2019} (CNN-rule) and that in \cite{Kim:2014} (CNN) on the data sets under study. Errors bars denote 95\% confidence intervals around the mean.}
\label{fig:01}
\end{figure}

\begin{figure}[!b]
\centering
\includegraphics[width = 0.9\textwidth]{images/legend.pdf}
\subfloat[SST2]{\includegraphics[width = 0.33\textwidth]{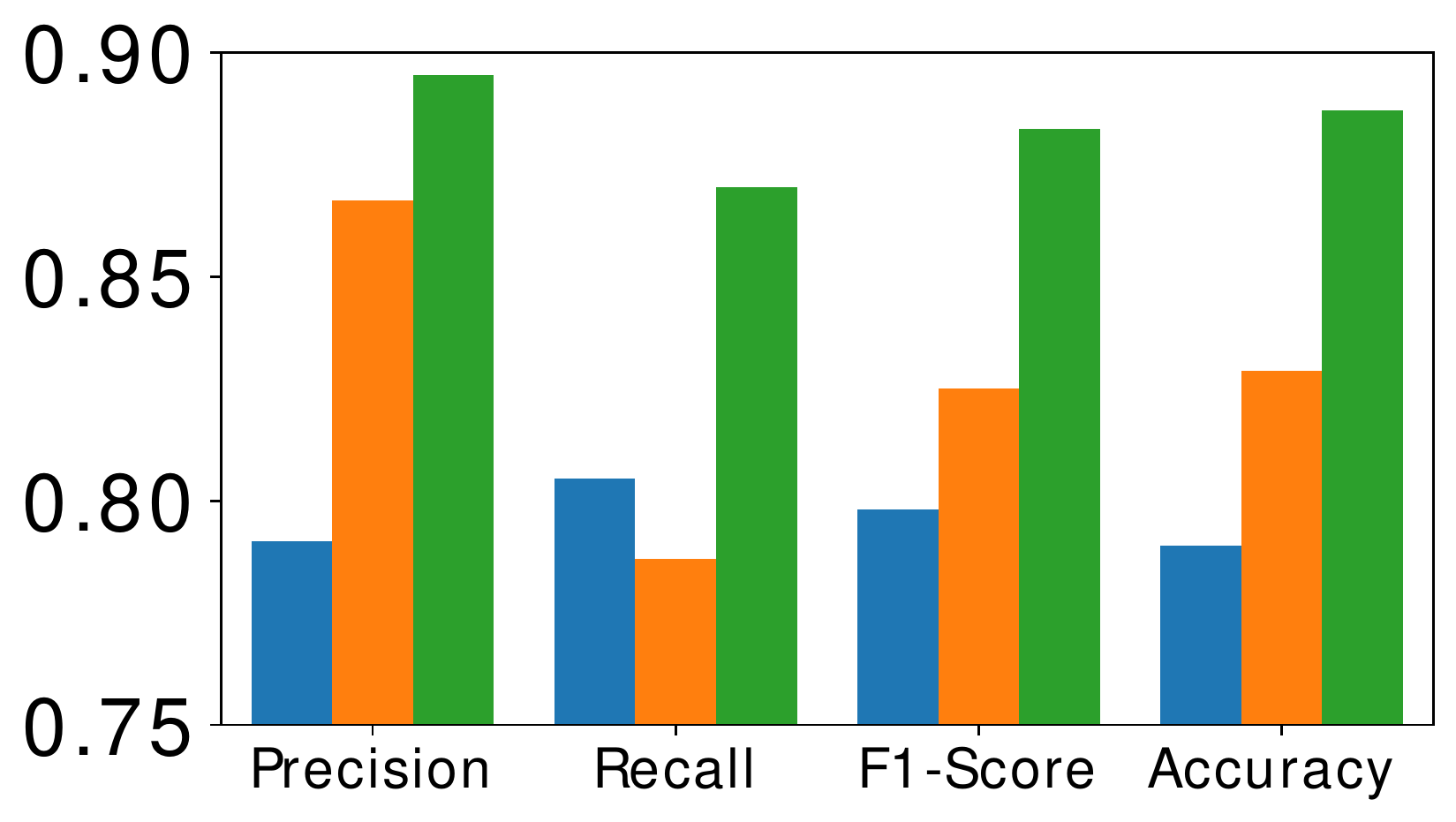}}
\subfloat[MR]{\includegraphics[width = 0.33\textwidth]{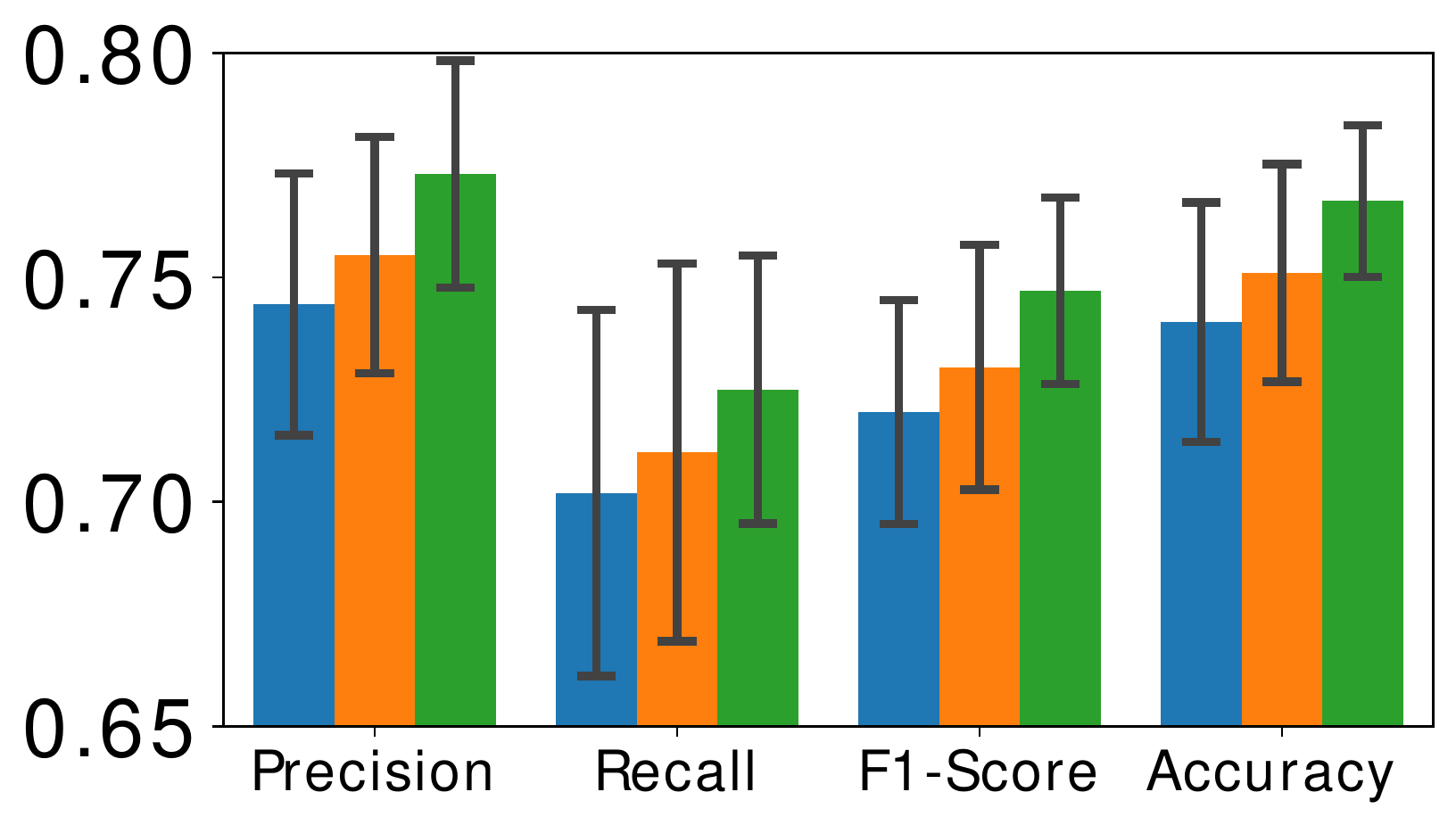}}
\subfloat[CR]{\includegraphics[width = 0.33\textwidth]{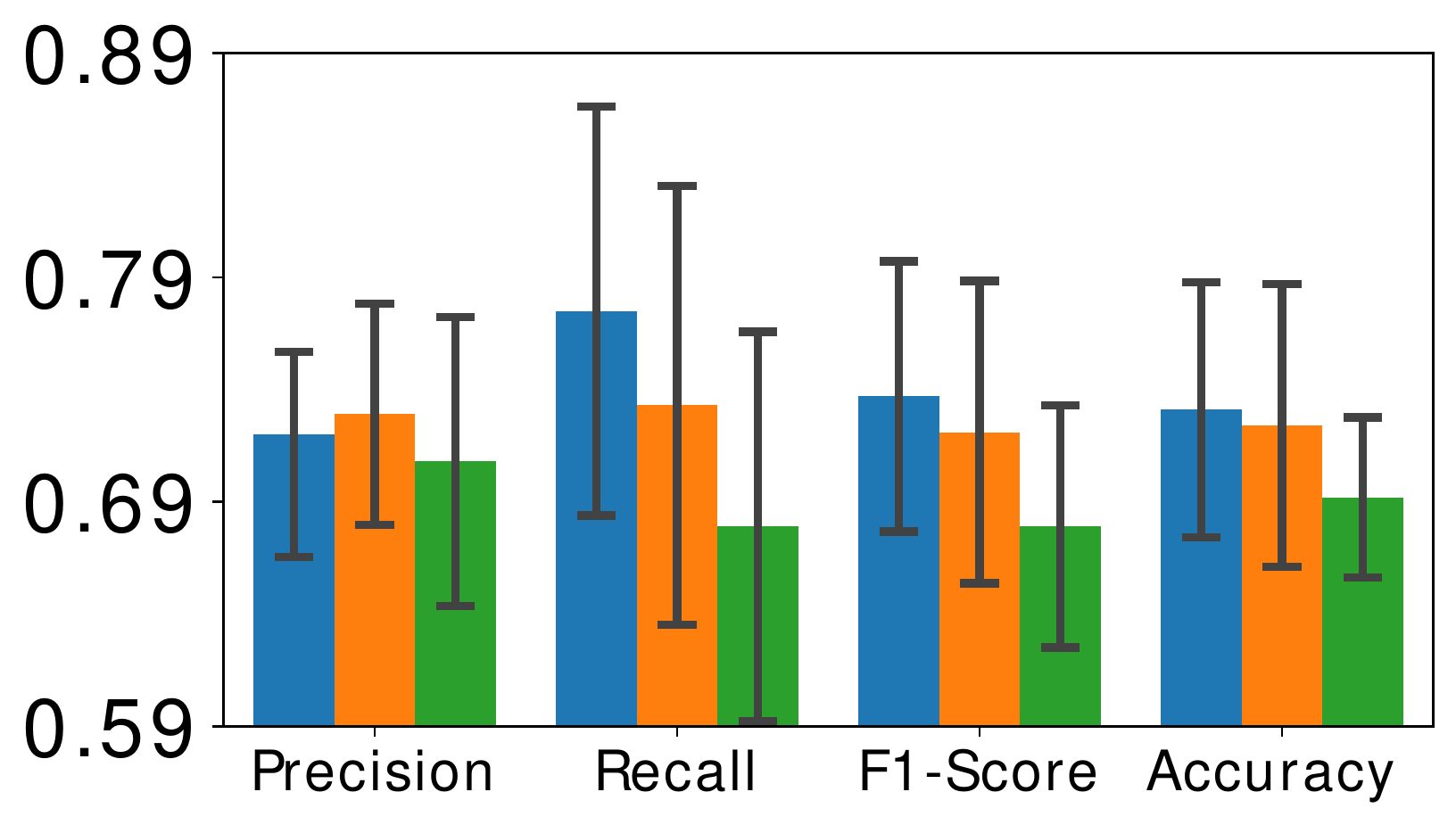}}
\caption{Performance obtained using our method (CNN-F), the method in \cite{Hu:2019} (CNN-rule) and that in \cite{Kim:2014} (CNN) on the data sets under study making use only of sentences containing \textbf{with A-but-B} structures. Errors bars denote 95\% confidence intervals around the mean.}
\label{fig:02}
\end{figure}

\begin{table}[!t]
    \caption{Performance obtained using our method (CNN-F), the method in \cite{Hu:2019} (CNN-rule) and that in \cite{Kim:2014} (CNN) on the data sets under study.}
    \centering
    \begin{tabular}{|*{5}{c|}}
    \hline
        \textbf{Method} &\multicolumn{4}{c|}{\textbf{SST2}}\\
        \hline
             &\textbf{Precision}  &\textbf{Recall}  &\textbf{F-1 Score}  &\textbf{Accuracy}\\
                 \hline
                    CNN &0.853 &\textbf{0.912} &0.881 &0.877\\
                    \hline
                    CNN-rule &0.878 &0.891 & 0.884 &0.884\\
                    \hline
                    CNN-F &\textbf{0.881} &0.895 &\textbf{ 0.888} &\textbf{0.887}\\
                    \hline
        &\multicolumn{4}{c|}{\textbf{MR}}\\
        \hline
             &\textbf{Precision}  &\textbf{Recall}  &\textbf{F-1 Score}  &\textbf{Accuracy}\\
                 \hline
                    CNN &0.826$\pm$0.012 &0.805$\pm$0.008 &0.815$\pm$0.005 &0.817$\pm$0.005\\
                    \hline
                    CNN-rule &0.826$\pm$0.012 &0.810$\pm$0.012 &0.818$\pm$0.008 &0.820$\pm$0.006\\
                    \hline
                    CNN-F &\textbf{0.830$\pm$0.009} &\textbf{0.810$\pm$0.007} &\textbf{0.818$\pm$0.004} &\textbf{0.820$\pm$0.004}\\
                    \hline
        &\multicolumn{4}{c|}{\textbf{CR}}\\
        \hline
             &\textbf{Precision}  &\textbf{Recall}  &\textbf{F-1 Score}  &\textbf{Accuracy}\\
                 \hline
                    CNN &0.881$\pm$0.020 &\textbf{0.881$\pm$0.018} &\textbf{0.880$\pm$0.012} &\textbf{0.847$\pm$0.014}\\
                    \hline
                    CNN-rule &\textbf{0.884$\pm$0.017} &0.869$\pm$0.020 &0.876$\pm$0.011 &0.844$\pm$0.012\\
                    \hline
                    CNN-F &0.879$\pm$0.016 &0.863$\pm$0.024 &0.870$\pm$0.013 &0.836$\pm$0.013\\
                    \hline
    \end{tabular}
    \label{tab:01}
\end{table}

\begin{table}[!t]
    \caption{Performance obtained using our method (CNN-F), the method in \cite{Hu:2019} (CNN-rule) and that in \cite{Kim:2014} (CNN) on the data sets under study making use only of sentences containing \textbf{with A-but-B} structures.}
    \centering
    \begin{tabular}{|*{5}{c|}}
    \hline
        \textbf{Method} &\multicolumn{4}{c|}{\textbf{SST2}}\\
        \hline
             &\textbf{Precision}  &\textbf{Recall}  &\textbf{F-1 Score}  &\textbf{Accuracy}\\
                 \hline
                    CNN &0.791 &0.805 &0.798 &0.790\\
                    \hline
                    CNN-rule &0.867 &0.787 &0.825 &0.829\\
                    \hline
                    CNN-F &\textbf{0.895} &\textbf{0.870} &\textbf{0.883} &\textbf{0.887}\\
                    \hline
        &\multicolumn{4}{c|}{\textbf{MR}}\\
        \hline
             &\textbf{Precision}  &\textbf{Recall}  &\textbf{F-1 Score}  &\textbf{Accuracy}\\
                 \hline
                    CNN &0.744$\pm$0.029 &0.702$\pm$0.041 &0.720$\pm$0.025 &0.740$\pm$0.027\\
                    \hline
                    CNN-rule &0.750$\pm$0.026 &0.711$\pm$0.042 &0.730$\pm$0.027 &0.751$\pm$0.024\\
                    \hline
                    CNN-F &\textbf{0.773$\pm$0.025} &\textbf{0.725$\pm$0.030} &\textbf{0.747$\pm$0.020} &\textbf{0.767$\pm$0.017}\\
                    \hline
        &\multicolumn{4}{c|}{\textbf{CR}}\\
        \hline
             &\textbf{Precision}  &\textbf{Recall}  &\textbf{F-1 Score}  &\textbf{Accuracy}\\
                 \hline
                    CNN &0.720$\pm$0.055 &\textbf{0.775$\pm$0.091} &\textbf{0.737$\pm$0.060} &\textbf{0.731$\pm$0.057}\\
                    \hline
                    CNN-rule &\textbf{0.729$\pm$0.049} &0.733$\pm$0.098 &0.721$\pm$0.067 &0.724$\pm$0.063\\
                    \hline
                    CNN-F &0.708$\pm$0.064 &0.679$\pm$0.087 &0.679$\pm$0.054 &0.692$\pm$0.036\\
                    \hline
    \end{tabular}
    \label{tab:02}
\end{table}

In Table \ref{tab:01} and Figure \ref{fig:01}, we show the precision, recall, F-1 score and accuracy for the positive sentiment class yielded by our method (CNN-F), the method in \cite{Hu:2019} (CNN-rule) and that in \cite{Kim:2014} (CNN).
From the experimental results, we observe that our method outperforms the two methods on both the SST2 and MR data sets for all measures. 
Note that the performance decreases on the CR data set for both CNN-rule and CNN-F, which indicates that the ``A-but-B'' rule cannot be generalized to data points coming from similar distributions.

In Table \ref{tab:02} and Figure \ref{fig:02}, we show the results  obtained only on sentences having the ``A-but-B'' syntactic structure.
At first glance, we note that our method works as intended and is quite competitive, outperforming the two baselines despite using only one rule for comparison. 
Since our method represents knowledge purely in terms of a distribution on input data, we can argue that it was bound to perform better than iterative-knowledge distillation \cite{Hu:2019} since the neural network will only process the input features that are consistent with the human knowledge. 
Also, a decrease in performance is observed on the CR data set for both CNN-rule and CNN-F, which is consistent with the fact that the ``A-but-B'' rule cannot be generalized for every sentence and should not be encoded in the parameters permanently.

Finally, the results in both tables
indicate that when there is a performance gain on datasets SST2 and MR by incorporating A-but-B rule, it is best for CNN-F and when there is a performance drop, it is worst for CNN-F. 
This suggests that the feature extracting functions not only can be used as an alternative
but also in conjunction with iterative knowledge distillation in order to provide a ``Maximum Performance Gain or Drop value'' from the constructed logic rules. 
This value can be used to select the best combination of these rules
via cross validation when using distillation. This also provides a mechanism to quantitatively evaluate how effectively the rule knowledge was distilled into the parameters of the neural network when applied to distillation approaches.

\section{Conclusion}
In this paper, we have shown how feature extracting functions can be employed to learn logic rules for sentiment analysis. This provides a means to representing human knowledge in neural networks via programmable feature extracting functions. 
Moreover, we have shown that, using these feature extracting functions, we can obtain a model whose posterior output can be influenced by domain knowledge expressed in terms of logic rules without the need of transferring these into the network parameters. The approach presented here is quite general in nature, being applicable to a wide variety of logic rules that can be expressed using rule-to-knowledge conditional probability distributions. We have illustrated the utility of our method for textual sentiment analysis and compared our results with those obtained using two baselines. In our experiments, our method was quite competitive, outperforming the alternatives.  

\bibliographystyle{splncs04}
\bibliography{egbib}

\begin{thebibliography}{10}
\providecommand{\url}[1]{\texttt{#1}}
\providecommand{\urlprefix}{URL }
\providecommand{\doi}[1]{https://doi.org/#1}

\bibitem{Bach:2015}
Bach, S.H., Broecheler, M., Huang, B., Getoor, L.: Hinge-loss markov random
  fields and probabilistic soft logic. CoRR  \textbf{abs/1505.04406} (2015)

\bibitem{Ratner:2017}
Bach, S.H., Rodriguez, D., Liu, Y., Luo, C., Shao, H., Xia, C., Sen, S.,
  Ratner, A., Hancock, B., Alborzi, H., Kuchhal, R., R\'{e}, C., Malkin, R.:
  Snorkel drybell: A case study in deploying weak supervision at industrial
  scale. In: Proceedings of the 2019 International Conference on Management of
  Data. p. 362–375 (2019)

\bibitem{Garcez:2002}
Gabbay, A., Garcez, A., Broda, K., Gabbay, D.M., Gabbay, P.: Neural-Symbolic
  Learning Systems: Foundations and Applications. Springer London (2002)

\bibitem{Ganchev:2010}
Ganchev, K., Gra\c{c}a, J.a., Gillenwater, J., Taskar, B.: Posterior
  regularization for structured latent variable models. Journal of Machine
  Learning Research  \textbf{11},  2001–2049 (2010)

\bibitem{Hinton:2015}
Hinton, G., Vinyals, O., Dean, J.: Distilling the knowledge in a neural
  network. In: NIPS Deep Learning and Representation Learning Workshop (2015)

\bibitem{Hu-and-Liu:2004}
Hu, M., Liu, B.: Mining and summarizing customer reviews. In: Proceedings of
  the Tenth ACM SIGKDD International Conference on Knowledge Discovery and Data
  Mining. pp. 168--177 (2004)

\bibitem{Hu:2019}
Hu, Z., Ma, X., Liu, Z., Hovy, E., Xing, E.: Harnessing deep neural networks
  with logic rules. In: Proceedings of the 54th Annual Meeting of the
  Association for Computational Linguistics (Volume 1: Long Papers). pp.
  2410--2420. Association for Computational Linguistics, Berlin, Germany (2016)

\bibitem{Kim:2014}
Kim, Y.: Convolutional neural networks for sentence classification. CoRR
  \textbf{abs/1408.5882} (2014)

\bibitem{krishna:2018}
Krishna, K., Jyothi, P., Iyyer, M.: Revisiting the importance of encoding logic
  rules in sentiment classification. In: Proceedings of the 2018 Conference on
  Empirical Methods in Natural Language Processing. pp. 4743--4751 (2018)

\bibitem{Mikolov:2014}
Le, Q.V., Mikolov, T.: Distributed representations of sentences and documents.
  CoRR  \textbf{abs/1405.4053} (2014)

\bibitem{lewis:92}
Lewis, D.D.: Feature selection and feature extraction for text categorization.
  In: Proceedings of the Workshop on Speech and Natural Language. pp. 212--217
  (1992)

\bibitem{Liang:2018}
Liang, X., Hu, Z., Zhang, H., Lin, L., Xing, E.P.: Symbolic graph reasoning
  meets convolutions. In: Advances in Neural Information Processing Systems.
  pp. 1853--1863 (2018)

\bibitem{Nguyen:2015}
Nguyen, A.M., Yosinski, J., Clune, J.: Deep neural networks are easily fooled:
  High confidence predictions for unrecognizable images. CoRR
  \textbf{abs/1412.1897} (2014)

\bibitem{pang-lee:2005}
Pang, B., Lee, L.: Seeing stars: Exploiting class relationships for sentiment
  categorization with respect to rating scales. In: Proceedings of the 43rd
  Annual Meeting of the Association for Computational Linguistics ({ACL}{'}05)
  (2005)

\bibitem{Socher:2014}
Pennington, J., Socher, R., Manning, C.: {G}love: Global vectors for word
  representation. In: Proceedings of the 2014 Conference on Empirical Methods
  in Natural Language Processing ({EMNLP}). pp. 1532--1543 (2014)

\bibitem{Peters:2018}
Peters, M., Neumann, M., Iyyer, M., Gardner, M., Clark, C., Lee, K.,
  Zettlemoyer, L.: Deep contextualized word representations. In: Proceedings of
  the 2018 Conference of the North {A}merican Chapter of the Association for
  Computational Linguistics: Human Language Technologies. pp. 2227--2237 (2018)

\bibitem{arun:2020}
Rai, A.: Explainable ai: from black box to glass box. Journal of the Academy of
  Marketing Science  \textbf{48} (2020)

\bibitem{Ribeiro:2016}
Ribeiro, M.T., Singh, S., Guestrin, C.: "why should {I} trust you?": Explaining
  the predictions of any classifier. CoRR  \textbf{abs/1602.04938} (2016)

\bibitem{Szegedy:2014}
Szegedy, C., Zaremba, W., Sutskever, I., Bruna, J., Erhan, D., Goodfellow, I.,
  Fergus, R.: Intriguing properties of neural networks. In: International
  Conference on Learning Representations (2014)

\bibitem{Tasker:2003}
Taskar, B., Guestrin, C., Koller, D.: Max-margin markov networks. In: NIPS. pp.
  25--32 (2003)

\bibitem{Tran:2017}
Tran, S.N.: Unsupervised neural-symbolic integration. CoRR
  \textbf{abs/1706.01991} (2017)

\bibitem{vilone:2020}
Vilone, G., Longo, L.: Explainable artificial intelligence: a systematic
  review. arXiv preprint  \textbf{arXiv:2006.00093} (2020)

\bibitem{Zeiler:2012}
Zeiler, M.D.: {ADADELTA:} an adaptive learning rate method. CoRR
  \textbf{abs/1212.5701} (2012)

\end{thebibliography}
\end{document}